\documentclass[conference]{IEEEtran}
\IEEEoverridecommandlockouts
\usepackage{cite}
\usepackage{amsmath,amssymb,amsfonts}
\usepackage{algorithmic}
\usepackage{graphicx}
\usepackage{textcomp}
\usepackage{xcolor}
\usepackage{multirow}
\usepackage{bm}
\usepackage{subfigure}

\def\BibTeX{{\rm B\kern-.05em{\sc i\kern-.025em b}\kern-.08em
    T\kern-.1667em\lower.7ex\hbox{E}\kern-.125emX}}
\begin{document}

\title{Transformer-Guided Convolutional Neural Network for Cross-View Geolocalization}


\author{\IEEEauthorblockN{Teng Wang}
\IEEEauthorblockA{\textit{School of Automation} \\
\textit{Southeast University}\\
wangteng@seu.edu.cn}
\and
\IEEEauthorblockN{Shujuan Fan}
\IEEEauthorblockA{\textit{School of Automation} \\
\textit{Southeast University}\\
fanshujuan@seu.edu.cn}
\and
\IEEEauthorblockN{Daikun Liu}
\IEEEauthorblockA{\textit{School of Automation} \\
\textit{Southeast University}\\
dkliu@seu.edu.cn}
\and
\IEEEauthorblockN{Changyin Sun}
\IEEEauthorblockA{\textit{School of Automation} \\
\textit{Southeast University}\\
cysun@seu.edu.cn}

}

\maketitle

\begin{abstract}
Ground-to-aerial geolocalization refers to localizing a ground-level query image by matching it to a reference database of geo-tagged aerial imagery. This is very challenging due to the huge perspective differences in visual appearances and geometric configurations between these two views. In this work, we propose a novel Transformer-guided convolutional neural network (TransGCNN) architecture, which couples CNN-based local features with Transformer-based global representations for enhanced representation learning. Specifically, our TransGCNN consists of a CNN backbone extracting feature map from an input image and a Transformer head modeling global context from the CNN map. In particular, our Transformer head acts as a spatial-aware importance generator to select salient CNN features as the final feature representation. Such a coupling procedure allows us to leverage a lightweight Transformer network to greatly enhance the discriminative capability of the embedded features. Furthermore, we design a dual-branch Transformer head network to combine image features from multi-scale windows in order to improve details of the global feature representation. Extensive experiments on popular benchmark datasets demonstrate that our model achieves top-1 accuracy of 94.12\% and 84.92\% on CVUSA and CVACT\_val, respectively, which outperforms the second-performing baseline with less than 50\% parameters and almost $\mathbf{2\times}$ higher frame rate, therefore achieving a preferable accuracy-efficiency tradeoff. 
\end{abstract}


\section{Introduction}

\begin{figure}[htbp]
   \centerline{\includegraphics{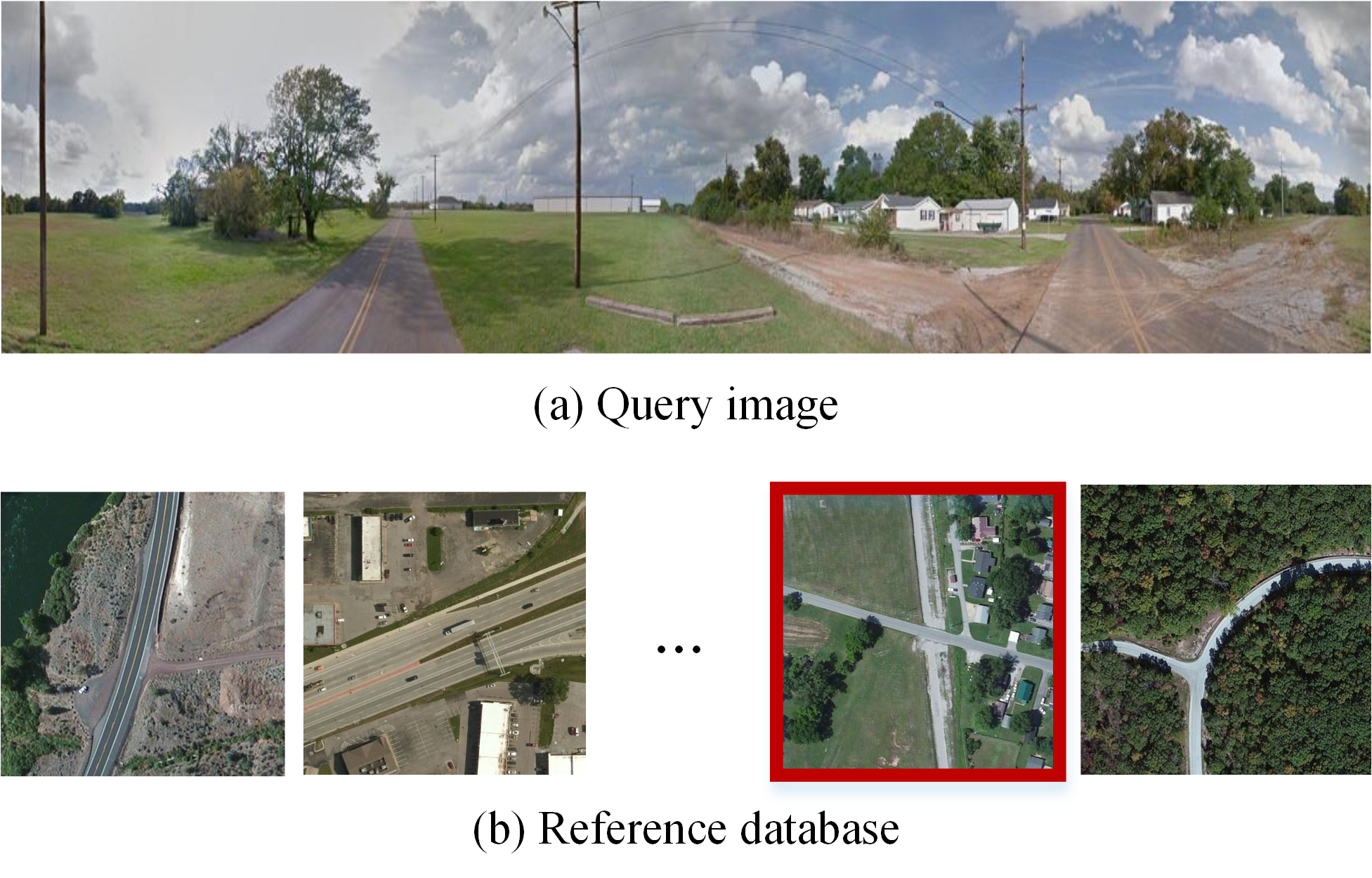}}
   \caption{Illustration of ground-to-aerial cross-view geolocalization task.}
    \label{fig1}
\end{figure}

Ground-to-aerial cross-view geolocalization aims to determine the geographical location of a query ground image by matching it with a large set of geo-referenced database aerial/satellite images, as illustrated in Fig.~\ref{fig1}. It plays an important role in robot navigation~\cite{Senlet2012}, autonomous driving~\cite{Mcmanus2014} and augmented reality~\cite{Middelberg2014}. Although it has broad applications, this task is extremely challenging due to the vast differences in both visual appearances and spatial configurations between the two views. As a consequence, learning effective feature representations for these two views can be very difficult and has thus attracted great attention of researchers. 

In the past few years, convolutional neural networks (CNNs) have significantly advanced the task of ground-to-aerial geolocalization. Prior works~\cite{Workman2015L, Workman2015W} extract features directly using the CNNs pre-trained for image classification. Later on, \cite{Lin2015, Vo2016, Vo2017} construct different Siamese CNNs to learn a more universal cross-view feature representation from corresponding datasets. Several more recent works incorporate CNNs with NetVlad layers~\cite{Hu2018}, capsule networks~\cite{Sun2019} or attention mechanisms~\cite{Cai2019, Shi2019} to further enhance discriminative feature representation learning. These CNN architectures typically leverage a stack of convolutional layers to increasingly enlarge receptive fields and collect local features in a hierarchical fashion as powerful feature representations. Despite the great improvements, these CNN-based models experience difficulty in capturing long-range dependency, and therefore fail to make more accurate predictions in complex scenarios, where visual signals are ambiguous or incomplete. 
 
Recently, Natural Language Processing (NLP) field has witnessed the rapid rise of Transformer with self-attention~\cite{Vaswani2017} in various powerful language modeling architectures, which model long-range interaction in a scalable and parameter-free manner. The success of Transformer on many NLP tasks strongly motivates researchers to push the limits of computer vision tasks by integrating CNN-based architectures with Transformer-style modules. With regard to cross-view geolocalization, Yang et al.~\cite{Yang2021} make the first attempt to introduce Transformer and  propose a novel structure, termed EgoTR, which consists of a ResNet backbone extracting CNN feature map from an input image and a Vision Transformer (ViT)~\cite{Dosovitskiy2021} modeling global context from the CNN map.  Such a Transformer-based model shows superiority over the dominant CNN-based counterparts. However, this structure requires a large amount of parameters and cannot meet real-time processing requirement, both of which are practical constraints and hamper its widespread adoption. 

\begin{figure*}[!htbp]
    \centerline{\includegraphics[width = 0.99\textwidth, height = 7.4cm]{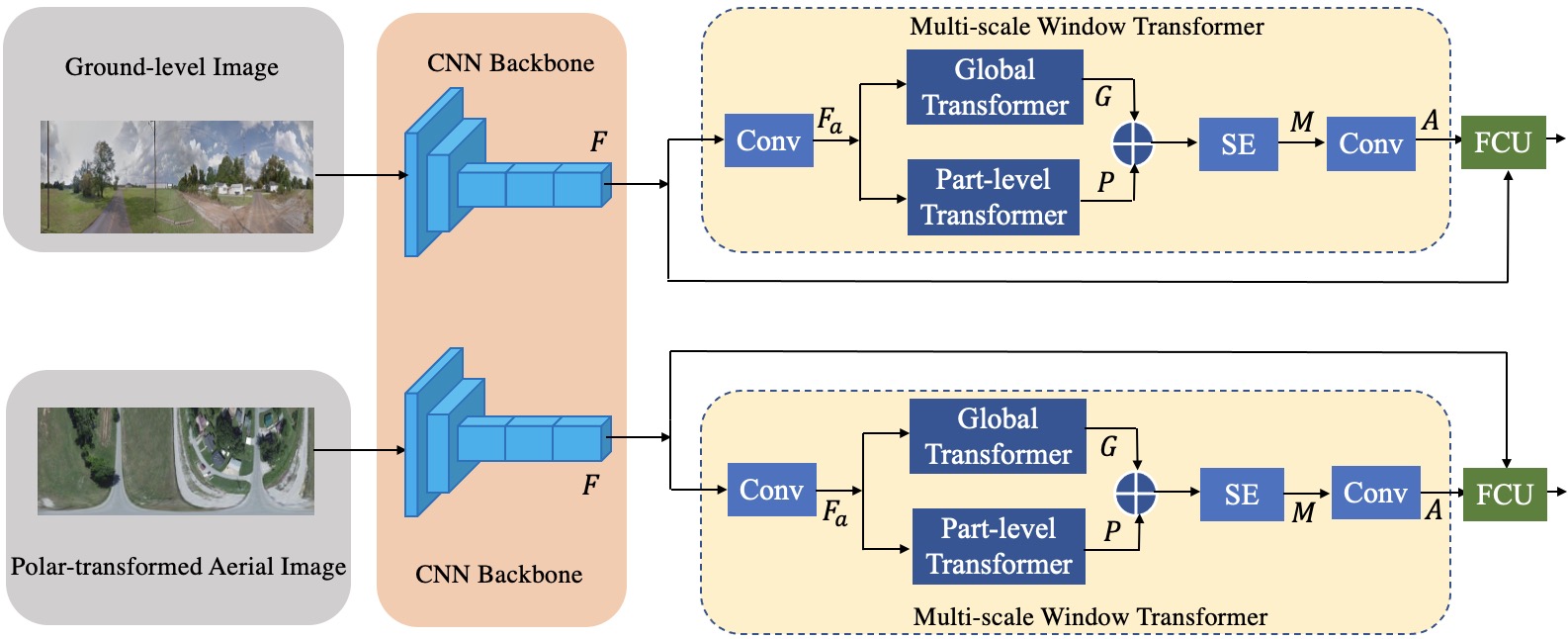}}
    \caption{Overview of our proposed Transformer-guided convolutional neural network architecture for cross-view geolocalization.}
    \label{fig:architecture}
\end{figure*}

In this paper, we hypothesize that Transformer could be strategically introduced into the CNN structure to improve accuracy of ground-to-aerial geolocalization, while concurrently maintaining a high degree of memory efficiency and frame rate. To verify our hypothesis, we propose a novel network structure, called \textbf{Trans}former-\textbf{G}uided \textbf{C}onvolutional \textbf{N}eural \textbf{N}etwork (TransGCNN), which couples CNN-based local features with Transformer-based global representations for enhanced representation learning. As shown in Fig.~\ref{fig:architecture}, TransGCNN consists of a CNN backbone and a Transformer head. The CNN backbone follows the design of VGG16~\cite{Simonyan2015} to extract local features, whereas the Transformer head operates on CNN feature map to model global context. Different from prior works, our Transformer head acts as a spatial attention module to select salient CNN features as the final feature representation. Such a coupling procedure allows us to leverage a lightweight Transformer network to greatly enhance the discriminative capability of the embedded features. Furthermore, considering that vanilla ViT block performs self-attention over the whole input map and easily miss local details, we design a dual-branch vision Transformer to combine image features from multi-scale windows to produce stronger features. Specifically, at a global-level, we perform self-attention over the whole input feature map; whereas at the part-level, we split the input feature map into several non-overlapping vertical parts and perform self-attention over each part separately. The enhanced feature maps from multi-scale windows are fused dynamically to complement each other.

We conduct extensive experiments on two popular benchmarks, CVUSA~\cite{Zhai2017} and CVACT\_val~\cite{Liu2019}. Experiment results demonstrate that our TransGCNN achieves state-of-the-art performance, while being lightweight and fast. It outperforms prior CNN-based models by a large margin in recalls. On the other hand, it achieves competitive recalls while utilizing fewer parameters and attaining higher frame rate compared to prior Transformer-based model. Main contributions of our work are summarized as follows:  

\begin{itemize}
\item We design a hybrid network structure, which leverages a Transformer-style head network as the spatial attention module to select salient CNN features as the final image descriptor. Such a coupling mechanism enables our model to improve the discriminative capability of the embedded features, while maintaining lightweight and fast.
  
\item We propose a novel multi-scale window Transformer by performing self-attention over multi-scale windows of the input map separately and fusing multi-scale information dynamically. This module has proved to enhance details of the global representation.

\item Extensive experiments on benchmark datasets confirm that our model achieves a preferable trade-off between accuracy and efficiency. It achieves competitive accuracy as the second-performing EgoTR  with less than 50\% parameters and about $\mathbf{2\times}$ higher frame rate.
 \end{itemize}

The rest of this paper is organized as follows. In Section~\ref{sec:related}, we provide an overview of related literature in cross-view geolocalization. Section~\ref{sec:framework} gives a detailed description of our proposed approach. In Section~\ref{sec:experiments}, we evaluate the performance of our approach by conducting extensive experiments for ablation studies and comparing it to the state-of-the-art methods. We draw conclusions in Section~\ref{sec:conclusion}.

\section{Related Work}
\label{sec:related}

\subsection{Traditional approaches}
In the early stage, handcrafted features are widely used to address the tasks of image matching and image retrieval. Hays et al.~\cite{Hays2008} design a variety of features by expert knowledge of designers, including color histogram, texton histograms, line features, gist descriptors, geometric context, and employ the $\ell_{2}$ distance between the above feature descriptors as the distance metric to search over a large-scale image database. Lin et al.~\cite{Lin2013} represent each image using four features, including HoG~\cite{Dalal2005}, self-similarity~\cite{Shechtman2007}, gist~\cite{Oliva2001}, and color histograms, and proposes two data-driven approaches to achieve cross-view geolocalization. In recent years, SIFT~\cite{lowe2004}, SURF~\cite{Bay2006}, ORB~\cite{Rublee2011} receive much attention due to the fact that they are invariant to rotation, scale scaling and brightness change, and also stable to a certain extent for angle change, affine transformation and noise, and are widely used in regular image matching task~\cite{Bastanlar2010, Huijuan2011, Sedaghat2015, Karami2017}. Inspired by the success of these features descriptors in regular image matching, \cite{Lefevre2017, Noda2010, Viswanathan2014} explore SIFT and SURF descriptors to match ground images against aerial images for cross-view geolocalization. However, these handcrafted features are less robust to lighting variations, viewpoint changes and other noises, thus limiting the use of these approaches in real applications to some extent.

\subsection{Deep learning based approaches}
Inspired by the great success of CNNs in many computer vision tasks, Workman and Jacobs~\cite{Workman2015L, Workman2015W} first introduce CNNs into the field of cross-view image retrieval, and employ the CNNs pre-trained for classification to extract deep features. The above works indicate better discriminativeness of deep features compared to handcrafted features.  However, these approaches directly utilize the generic feature representations designed for object classification, resulting in relatively low image retrieval accuracy. To that end, Lin et al.~\cite{Lin2015} construct a two-branch Siamese network to map two views into a common feature embedding space, and a weighted soft-margin triplet loss is utilized to speed up the network learning; VO and Hays~\cite{Vo2016} explore several deep CNN architectures including Classification, Hybrid, Siamese and Triplet networks for cross-view matching, and experimental results prove that triplet network outperforms other CNN architectures by a large margin. Recent works put more emphasis on designing CNN architectures for better representation learning. Hu et al.~\cite{Hu2018} incorporate a learnable NetVLAD layer~\cite{Arandjelovic2016} into a two-branch Siamese CNN architecture to learn view-invariant feature representations. Sun et al.~\cite{Sun2019} couple the powerful ResNet~\cite{He2016} backbone network with capsule layers~\cite{Sabour2017} to model high-level semantics. Cai et al.~\cite{Cai2019} propose a lightweight attention module by combining spatial and channel attention mechanisms to emphasis visually salient features in input images. Shi et al.~\cite{Shi2019} design a multi-head spatial attention module to alleviate content-dependent domain discrepancy. Despite great improvements, few of the above works consider the long-distance relationships among visual elements, which decreases the discriminative capability of the embedded features. To tackle the challenge, Yang et al.~\cite{Yang2021} make the first exploration to introduce transformer and propose a hybrid structure consisting of a ResNet backbone extracting CNN feature map from an input image and a ViT modeling global context from the CNN map.  However, this hybrid structure requires a large amount of parameters, and suffers from low frame rate. Different from existing methods, we design a novel network structure, which leverages multi-scale window Transformer as the spatial attention module to select salient CNN features as the final image descriptor. We demonstrate that our method achieves a preferable trade-off between accuracy and efficiency.

\section{The Proposed Method}
\label{sec:framework}

\subsection{Overview}
In this work, the ground-to-aerial geolocalization problem is regarded as a standard image retrieval task. We follow~\cite{Shi2019} to employ a two-step approach to solve it. The first step is to apply a regular polar transform to warp aerial images in order to build more apparent spatial correspondences between aerial and ground view images. Then, we adopt a domain-specific Siamese-like architecture with two independent branches of the same architecture to separately learn representations for these two views. A novel Transformer-guided convolutional neural network, termed TransGCNN, is proposed for each branch. The design of TransGCNN draws inspiration from how human beings observe the world. Given a scene, humans typically first glance at the global view and then gaze into salient local details under the guidances of the global context to recognize the scene. Motivated by this, we propose TransGCNN,  which consists of a CNN backbone, where the first thirteen layers of  VGG16~\cite{Simonyan2015} are adopted to capture the local patterns, and a multi-scale window Transformer head, where self-attention is performed over multi-scale windows of the CNN feature map to capture the global representation, as shown in Fig.~\ref{fig:architecture}. Given local and global features, we feed them into the feature coupling unit (FCU), which innovatively treats each channel of the global feature map as one spatial attention map to re-weight the CNN feature map for emphasizing salient local features.  We concatenate the embedded features from multiple attention channels together as our final image descriptor. During training, the weighted triplet loss is utilized to supervise the whole deep model.

\subsection{Polar Transform}
Based on the fact that pixels lying on the same azimuth direction in an aerial image approximately correspond to a vertical image column in the ground view image, we explicitly apply a regular polar transform to warp the aerial images and thus to some extent bridge the geometric correspondence gap between these two different view domains. Specifically, we take the center of each aerial image as the polar origin and the north direction as angle $0^{\circ}$ in the polar transform. Suppose the sizes of the original ground and aerial images are  $W_{g} \times H_{g}$ and $W_{a} \times H_{a}$, respectively. Given any point \(({x^s},{y^s})\) in an original aerial image, its new location \(({x^t},{y^t})\)  in the target transformed aerial image could be computed as follows.  

\begin{equation}
	\begin{split}
        {x^t} = \frac{W_{a}}{2} + \frac{A}{2}\frac{{{y^s}}}{H_{g}}\sin (\frac{{2\pi }}{W_{g}}{x^s})\\
        {y^t} = \frac{H_{a}}{2}  - \frac{A}{2}\frac{{{y^s}}}{H_{g}}\cos (\frac{{2\pi }}{W_{g}}{x^s})
    \end{split}
    \label{eq1}
\end{equation}
After polar transform, the objects in the transformed aerial images will lie in similar positions to their counterparts in the ground images, as shown in Fig.~\ref{fig:architecture}. Note that the size of the transformed satellite images is constrained to be the same as that of the ground ones in order to facilitate training of our two-branch Siamese network. More details about the polar transform could be found in~\cite{Shi2019}.

\subsection{Multi-Scale Window Transformer}
Our multi-scale window Transformer consists of two parallel branches: a global Transformer and a part-level Transformer. Both branches receives the CNN feature map $F_a$ as inputs. Here,  $F_a$ is obtained by performing $1\times1$ convolution on the CNN feature map $F$ to adjust its channel dimension, such that pre-trained parameters on ImageNet could be used to initialize these two Transformers. The enhanced feature maps from these two branches are fused to derive the final global representation.

\textbf{Global Transformer.}  This module contains $L$ repeated vision Transformer blocks~\cite{Dosovitskiy2021}. Each block consists of a single-head self-attention (SHSA) module and a multi-layer perceptron (MLP) block.  Layer normalization (LN) is applied before each layer and residual connection. To capture global context, we feed the CNN feature map $F_a \in \mathbb{R}^{H \times W \times C}$ into the $L$-layer Transformer encoder by reshaping it into a sequence of tokens $F_t \in \mathbb{R}^{HW \times C}$. Each Transformer block processes input features as follows:
\begin{equation}
\mathbf{z}_{\ell}' = \text{SHSA}(\text{LN}(\mathbf{z}_{\ell-1})) + \mathbf{z}_{\ell-1},
\end{equation}
\begin{equation}
\mathbf{z}_{\ell} = \text{MLP}(\text{LN}(\mathbf{z}_\ell')) + \mathbf{z}_{\ell}', 
\end{equation}
where $\mathbf{z}_{\ell} \in\mathbb{R}^{HW\times C}, \ell \in \{1,\cdots, L\}$, is the encoded image representations at the $\ell$-th block, and we set $\mathbf{z}_{\ell-1} = F_t$ for $\ell = 1$. SHSA enables the Transformer to model a global relationship in a parameter-free manner, which is formulated as:
\begin{equation}
\text{SHSA}(X)  = \text{Softmax}(\frac{QK^T}{\sqrt{D}})V. \hspace{0.4cm}
\end{equation}
Here, $Q$, $K$, $V\in\mathbb{R}^{N\times D}$ are respectively the $query$, $key$ and $value$ matrices, which are linear projections of input $X\in\mathbb{R}^{N\times D}$, $N$ is the length of input token sequence, and $D$ is the token embedding size. More details about vision Transformer block could be found in~\cite{Dosovitskiy2021}.
 
\begin{figure}[htbp]
    \centerline{\includegraphics[width=0.46\textwidth, height = 2.450in]{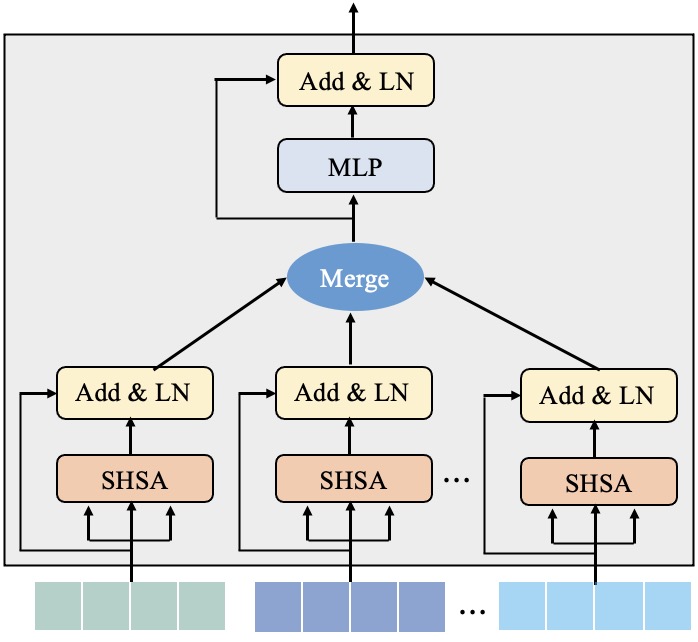}}
    \caption{An illustration of our part-level Transformer block.}
    \label{fig:mwtransformer}
\end{figure}

\textbf{Part-level Transformer.} Although the global Transformer could effectively capture long-range dependency, it easily misses local details. To this end, we introduce a part-level Transformer module to enhance local details of the global representation. This module is composed of $L$ part-level Transformer blocks, whose overall architecture is illustrated in Fig.~\ref{fig:mwtransformer}. The main difference between our proposed part-level Transformer block and the regular one is that we split the input feature map into multiple non-overlapping partitions and perform self-attention with shared parameters over each partition separately. As different scene parts along horizontal direction typically carry different semantics, we split the input feature map into $w$ vertical partitions. This process could be described as follows:
\begin{equation}
\mathbf{\hat{z}}_{\ell-1}= [\mathbf{\hat{z}}_{\ell-1}^{1,1}, \mathbf{\hat{z}}_{\ell-1}^{1,2}, \cdots, \mathbf{\hat{z}}_{\ell-1}^{H, W}],
\end{equation}
where $\mathbf{\hat{z}}_{\ell-1} \in\mathbb{R}^{HW\times C}$ is the encoded image representations at the $\ell-1$-th block, and we set $\mathbf{\hat{z}}_{\ell-1} = F_t$ for $\ell = 1$. $\mathbf{\hat{z}}_{\ell-1}^{i, j}$ is the feature token at position $(i, j)$ if reshaping the token sequence $\mathbf{\hat{z}}_{\ell-1}$ back into a 2D feature map. To capture local details,  $\mathbf{\hat{z}}_{\ell-1}$ is first split into $w$ partitions before being input into the $\ell$-th block:
\begin{equation}
\mathbf{\hat{z}}_{\ell-1}= [\mathbf{\hat{z}}_{\ell-1}^{1}, \mathbf{\hat{z}}_{\ell-1}^{2}, \cdots, \mathbf{\hat{z}}_{\ell-1}^{w}]
\end{equation}
where
\begin{equation}
\mathbf{\hat{z}}_{\ell-1}^{j} = [\mathbf{\hat{z}}_{\ell-1}^{1, I_j+1}, \cdots, \mathbf{\hat{z}}_{\ell-1}^{1, I_j+M}, \cdots, \mathbf{\hat{z}}_{\ell-1}^{H, I_j+1}, \cdots, \mathbf{\hat{z}}_{\ell-1}^{H, I_j+M}]
\end{equation}
where $M = \frac{W}{w}$, and $I_j = M*(j-1)$. Afterwards, we apply SHSA to each partition $\mathbf{z}_{\ell-1}^{j}\in \mathbb{R}^{HM \times C}$:
\begin{equation}
\mathbf{\hat{z}}_{\ell}'^{, j} = \text{SHSA}(\text{LN}(\mathbf{\hat{z}}_{\ell-1}^{j})) + \mathbf{\hat{z}}_{\ell-1}^{j}.
\end{equation}
And then, all partitions are merged back into one feature map and go through the remaining MLP module to fuse information across channels for each point separately: 
\begin{equation}
\mathbf{\hat{z}}_{\ell}'= \text{Merge}(\mathbf{\hat{z}}_{\ell}'^{,1}, \mathbf{\hat{z}}_{\ell}'^{,2}, \cdots, \mathbf{\hat{z}}_{\ell}'^{,w})
\end{equation}
\begin{equation}
\mathbf{\hat{z}}_{\ell} = \text{MLP}(\text{LN}(\mathbf{\hat{z}}_\ell')) + \mathbf{\hat{z}}_{\ell}'.
\end{equation}
where Merge is the operation to re-arrange points in each partition back in their original orders in the feature map. Such Split-Merge designs give part-level Transformer the advantage of retaining details while modeling long-range dependency. 

\begin{figure*}[htbp]
	\centering
	\begin{minipage}{0.48\linewidth}
		\centerline{\includegraphics[width=\textwidth]{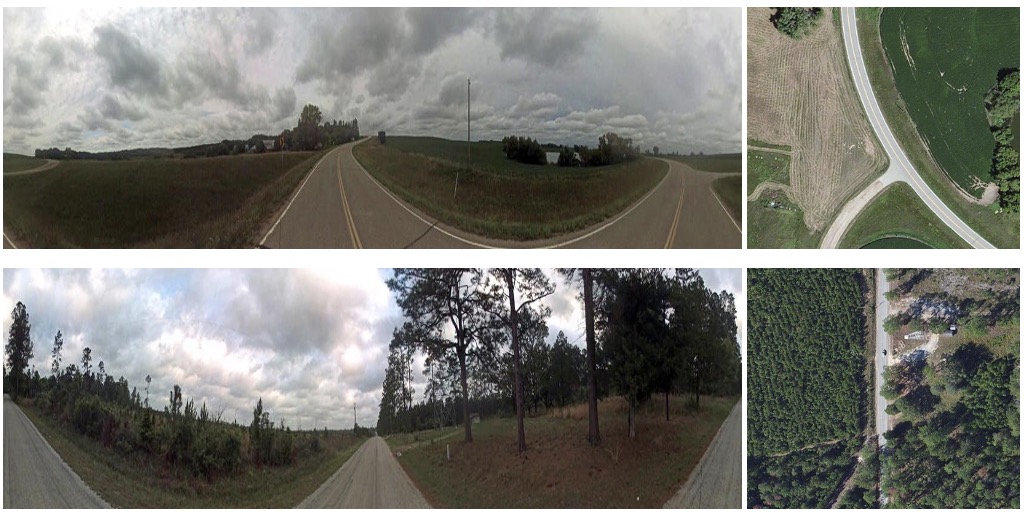}}
		\centerline{(a) CVUSA }
	\end{minipage}
	\begin{minipage}{0.48\linewidth}
		\centerline{\includegraphics[width=\textwidth]{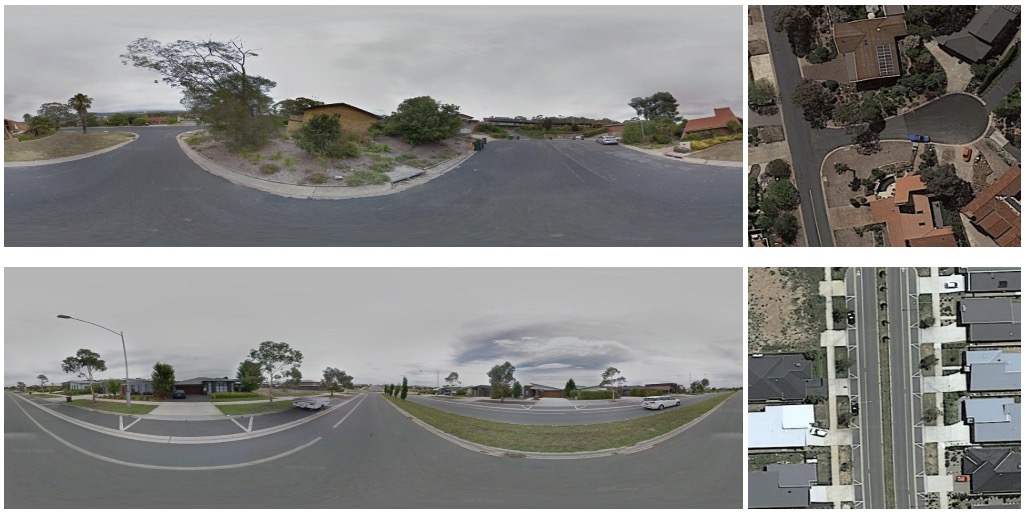}}
		\centerline{(b) CVACT}
	\end{minipage}
	\caption{Ground-to-aerial image pairs sampled from CVUSA and CVACT. Each subfigure illustrate a ground image (left) and an aerial image (right)}
	\label{fig_datasample}	
\end{figure*}

The output feature maps $G = \mathbf{z}_L$ and $P = \mathbf{\hat{z}}_{L}$ respectively from the global and part-level Transformers are fused by the element-wise summation followed by one SE (i.e., Squeeze-and-Excitation) block~\cite{Hu2018SENet} to derive the final global representation $M$.  Here, the SE block is introduced to capture channel-wise dependency and emphasize informative channels. It is worth mentioning that we do not utilize position encodings in both global and part-level Transformers. This is because the global feature map $M$ is just treated as the spatial attention map to select salient CNN features as the final embedded features, and the CNN features already carry abundant position information. 
    
\subsection{Feature Coupling Unit}
Given the local features $F$ and global representation $M$ respectively from the CNN backbone and Transformer head, how to utilize them to effectively enhance the discriminative capability of the embedded features is an important issue. To solve, we propose an attention-based feature coupling unit, where the global feature map $M$ is treated as the spatial attention map to guide the model to focus on salient features on the feature map $F$. Furthermore, motivated by the feature aggregation strategy~\cite{Shi2019}, we aim to improve the robustness of our feature representations by aggregating multiple embedded features. Towards this goal, we feed the global feature map $M$ into a $1\times 1$ convolution layer to adjust its channel dimension and treat each channel as one specific spatial attention map to generate the embedded features. The attention-based feature coupling procedure could be described as follows.
\begin{equation}
A = \mathbf{Conv_{1\times 1}}( M)
\end{equation}
\begin{equation}
f^{c}_{i} = <F^{c}, A^{i}>
\end{equation}
where $F^{c}$ represents the CNN feature map in the $c$-th channel, $A^{i}$ denotes the attention map in the $i$-th channel, $<\cdot, \cdot>$ denote the Frobenius product of the two inputs, and $f^{c}_{i}$ is the embedded feature activation for the $c$-th channel from the $i$-th attention branch. 

We concatenate the embedded features from multiple attention channels together as our final image descriptor $\textbf{f}$.
\begin{equation}
\textbf{f} = \{f^{1}_{1}, \cdots, f^{C}_{1}, f^{1}_{2}, \cdots, f^{C}_{2}, \cdots, f^{1}_{K}, \cdots, f^{C}_{K}\}
\end{equation}
where $C$ and $K$ represent channel dimensions of the CNN feature map $F$ and attention map $A$, respectively.

\subsection{Loss Function}
\label{subsec:aligned_loss}

Recently, triplet loss  is widely used in cross-view image matching and retrieval tasks~\cite{Shi2019, Hu2018, Liu2019}, and has proved to achieve impressive performances. It aims to minimize the distance between matched pairs while maximizing the distance between unmatched pairs in the embedding space. Similar to~\cite{Shi2019}, we employ a weighted soft-margin triplet loss to supervise the whole model. Specifically, given an image triplet $\left \langle I_{a}, I_{p}, I_{n} \right \rangle$, the retrieval loss $ L$ is defined as follows:
\begin{equation}
    \begin{split}
        L = \log (1 + {e^{{\rm{\gamma}}({d_{f}^{pos}} - {d_{f}^{neg}})}}),
    \end{split}
    \label{eq:globalloss}
\end{equation}
where $d_{f}^{pos}$ and $d_{f}^{neg}$ refer to the $\ell_{2}$  distance between the feature representations from matched $\left \langle I_{a}, I_{p} \right \rangle$ and unmatched  $\left \langle I_{a}, I_{n} \right \rangle$ image pairs, respectively; $\gamma$ is a parameter to adjust the gradient of the loss, and is empirically set to 10. 

\section{Experiments}
\label{sec:experiments}

\subsection{Experiment Details}
\textbf{Datasets.} We conduct all our experiments on two standard benchmark datasets: CVUSA~\cite{Zhai2017} and CVACT~\cite{Liu2019}. Each dataset provides a total of 35,532 ground-to-aerial image pairs for training.  CVUSA provides a total of 8,884 image pairs for testing and CVACT provides the same number of image pairs for validation (denoted as CVACT\_val). Additionally, CVACT also provides a total of 92,802 image pairs with accurate geo-tags to support fine-grained city-scale geo-localization (denoted as CVACT\_test). In CVACT\_test, all aerial images within 5 meters away from a query ground image are regarded as the ground-truth correspondences for the query image. That means there may exist several corresponding aerial images for each query ground image in this dataset. In this work, we follow previous works~\cite{Shi2019, Wang2021} to employ CVUSA and CVACT\_val to evaluate the performance of our proposed method on standard cross-view geolocalization task. In both datasets, each image pair consist of a panoramic street-view image and its ground-truth satellite image, which are both north-aligned.  Note that the street-view and aerial images in both datasets are captured at different times, making ground-to-aerial geolocalization more challenging. Sampled image pairs from these two datasets are presented in Fig.~\ref{fig_datasample}.

\begin{table*}[htbp]
   \renewcommand{\arraystretch}{1.7}
    \caption{Performance comparison on recall accuracy with the state-of-the-art on CVUSA \& CVACT\_val dataset.}
    \begin{center}
    \begin{tabular}{c|c|c|c|c|c|c|c|c}
    \hline
    \multirow{2}*{\textbf{Model} }    &  \multicolumn{4} {|c|} {\textbf{CVUSA}} &  \multicolumn{4} {|c} {\textbf{CVACT\_val}}\\
    \cline{2-9} 
    ~   &  \hspace{0.20cm}\textbf{r@1} \hspace{0.20cm}   & \hspace{0.20cm}\textbf{r@5} \hspace{0.20cm}   & \hspace{0.15cm}\textbf{r@10} \hspace{0.20cm}  & \hspace{0.15cm}\textbf{r@1\%}\hspace{0.20cm} &  \hspace{0.20cm}\textbf{r@1} \hspace{0.20cm}   & \hspace{0.20cm}\textbf{r@5} \hspace{0.20cm}   & \hspace{0.20cm}\textbf{r@10} \hspace{0.20cm}  & \hspace{0.20cm}\textbf{r@1\%}\hspace{0.20cm} \\
    \hline
    CVM-NET~\cite{Hu2018}                      &  22.53                         & 50.01                          & 63.19                           & 93.62                      & 20.15                   & 45.00                        & 56.87.                       & 87.57 \\
    
    Liu and Li~\cite{Liu2019}                      &  40.79                          & 66.82                          & 76.36                           & 96.08                      & 46.96                  & 68.28                        & 75.48                         & 92.01\\
    
    Regmi and Shah~\cite{Regmi2019}     &  48.75                          & -                                  & 81.27                           & 95.98                      & -                          & -                                & -                                 & -\\
  
    CVFT~\cite{Shi2020}                            &  61.43                          & 84.69                          & 90.49                           & 99.02                      & 61.05                  & 81.33                        & 86.52                         & 95.93\\
 
    SAFA~\cite{Shi2019}                            &  89.84                          & 96.93                          & 98.14                           & 99.64                      & 81.03                  & 92.80                        & 94.84                         & 98.17\\
  
    Shi et al.~\cite{Shi2020CVPR}             &  91.93                          & 97.50                          & 98.54                           & 99.67                      & 82.49                  & 92.44                        & 93.99                         & 97.32\\
    \hline 

    EgoTR ~\cite{Yang2021}                     &  93.69                          & 98.15                           & 98.90                          & 99.74                      & 84.19                  & 94.37                        & 95.85                         & 98.26 \\
    \hline 
  
    TransGCNN  (ours)                              & \textbf{94.15}              &\textbf{98.21 }               & \textbf{98.94}              & \textbf{99.79}          & \textbf{84.92}     &\textbf{94.46}              &\textbf{95.88}              &\textbf{98.36} \\
    \hline
    \end{tabular}
    \label{table:accuracy}
    \end{center}
    \end{table*}

\textbf{Implementation Details.} We resize both the panoramic street-view image and polar-transformed satellite image into $128 \times 512$, and perform a mean normalization on resized images before feeding them into our proposed TransGCNN. The first 13 layers of VGG16 with pretrained weights on ImageNet is employed as our CNN backbone to extract local features. The output CNN feature map is fed into a $1\times 1$ convolutional layer to increase its channel dimension to 768. The enhanced CNN feature map goes through the multi-scale window transformer module to capture the global representation. For part-level Transformer, we divide the input CNN feature map into 5 vertical partitions and perform self-attention over each part separately. We empirically set the depth $L$ of both global and part-level Transformer to 2 and initialize them with pretrained parameters on ImageNet. The channel dimension $K$ of the attention map is set to $8$. Our experiments are based on Pytorch. The models are trained using AdamW optimizer with an initial learning rate set to $10^{-5}$. If not specified, all the experiments are conducted on a 32GB NVIDIA GeForce RTX 3090 GPU, and the batch size $B$ is set to 32. We utilize exhaustive mini-batch strategy~\cite{Vo2016}  to generate image triplets within a batch. For each ground-view image in a mini-batch, there are 1 matching aerial images and  $B-1$ unmatched aerial images, generating a total of $B(B-1)$ triplets. Similarly, there are 1 matching ground-view images and $B-1$ unmatched ground-view images for each satellite image, creating $B(B-1)$ triplets in total. Thus, we have a total of $2B(B-1)$ image triplets within each mini-batch.

\textbf{Evaluation Metrics.} We adopt the recall accuracy at top K  as the metric to evaluate the ground-to-aerial image retrieval accuracy of our deep model. In this work, we follow~\cite{Shi2019} to consider the recall accuracy at top-1, top-5, top-10, up to  top-1\%   (i.e., r@1, r@5, r@10, r@1\%). Besides, we count the number of parameters and frames per second (FPS) to measure model size and processing speed of our method, respectively.  

\subsection{Comparison to State-of-the-Art Models}
In this section, we conduct extensive evaluations on the standard benchmarks CVUSA~\cite{Zhai2017} and CVACT\_val~\cite{Liu2019} by comparing our TransGCNN with several recent state-of-the-art methods: CVM-Net~\cite{Hu2018}, Liu and Li~\cite{Liu2019}, CVFT\cite{Shi2020}, SAFA~\cite{Shi2019}, Shi et al.~\cite{Shi2020CVPR}, and EgoTR~\cite{Yang2021}. Among these six baseline models,  the first five models employ stacked convolutional layers to collect local features as the final feature representation, while the last EgoTR  introduces vision Transformer ViT to capture the global representation.  It is worth mentioning that our TransGCNN shares similar structure with CNN-based SAFA, except that we leverage multi-scale window Transformer to design the spatial attention module for emphasizing salient local features. For fair performance comparison, we compute the recall accuracies of these existing approaches using the released models or codes provided by the authors.    

\begin{figure*}[htbp]
	\centering
	\begin{minipage}{0.495\linewidth}
		\centerline{\includegraphics[width=\textwidth, height = 2.5in]{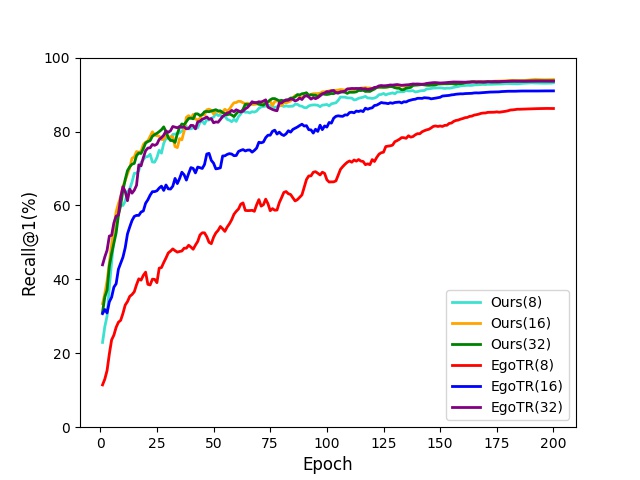}}
		\centerline{(a) Learning curves }
	\end{minipage}
	\begin{minipage}{0.495\linewidth}
		\centerline{\includegraphics[width=\textwidth, height = 2.5in]{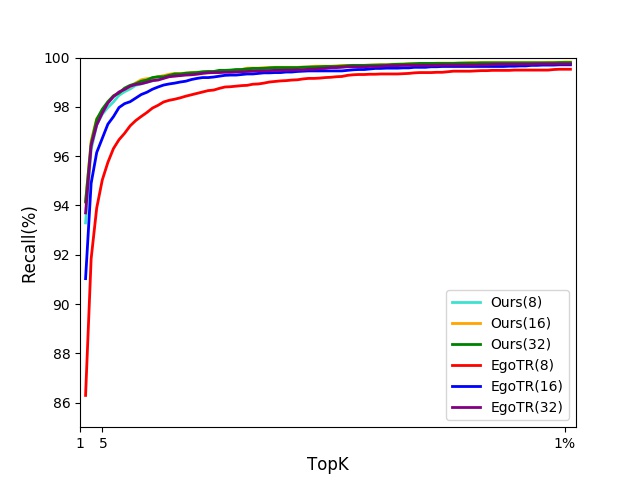}}
		\centerline{(b) Recalls}
	\end{minipage}
	\caption{Comparison of learning curves and recalls of different models on CVUSA datasets.}
	\label{fig_learning}	
\end{figure*}

\textbf{Accuracy.} We evaluate our TransGCNN on standard cross-view geolocalization. Table~\ref{table:accuracy} reports recalls at top-1, top-5, top-10, up to top 1\%  on CVUSA and CVCAT\_val datasets. From the experiment results, we observe that our TransGCNN significantly surpasses those CNN-based baselines in the most challenging  r@1 accuracy, indicating better discriminative capability of the feature representation learned from our TransGCNN. In particular, our TransGCNN achieves r@1 of 94.15\% on CVUSA dataset compared to 89.84\% obtained by SAFA, and the TransGCNN outperforms SAFA by a significant margin of 3.89\% points at r@1 on CVACT\_val dataset. The superiority of our proposed TransGCNN over CNN-based SAFA suggests the effectiveness of leveraging Transformer-style module to implement the spatial attention mechanism. This is because Transformer excels at capturing global contexts, which are typically critical for recognizing discriminative parts.  In addition, our TransGCNN performs slightly better than the Transformer-based EgoTR on both datasets (by 0.43\% on CVUSA and by 0.73\% on CVACT\_val). In summary, our TransCGNN achieves 94.15\% and 84.92\% top-1 accuracy on CVUSA and CVCAT\_val datasets, respectively, which are new state-of-the-art performances.

\textbf{Model Size and Processing Speed.} Real-world ground-to-aerial geolocalization applications also pose high requirements on model size and processing speed of the cross-view image retrieval network. Thus, we count the amount of parameters for different models, and test these models with identical testing settings to record their FPS. Specifically, we run all these models on a RTX 3090, 5 Core Intel(R) Xeon(R) Silver 4210R CPU@2.40GHz. The results are listed in Table~\ref{table:speed}. We observe that our TransGCNN requires $2.23\times$ fewer parameters and attains almost $2\times$ higher FPS compared to EgoTR, while achieving higher recalls on both datasets. This observation could be explained as follows. Our TransGCNN employ only 2 Transformer blocks in both global and part-level Transformers to design the spatial attention module, however EgoTR stacks 12 Transformer blocks to design the backbone network, which greatly increases the number of parameters and processing time per frame. In addition, compared to CNN-based SAFA, our TransGCNN improve r@1 accuracy by a significant margin (i.e., 4.31\% and 3.89\% respectively for CVUSA and CVACT\_Val) at the cost of requiring $2.97\times$ more parameters and obtaining $1.55\times$ lower FPS. The above two observations demonstrate that our TransGCNN significantly improves the recall accuracy of dominant CNN-based solutions, while maintaining a relatively higher memory efficiency and frame rate compared to the second-performing EgoTR. In other words, our TransGCNN achieves a preferable trade-off between accuracy and efficiency.

\begin{table}[htbp]
\renewcommand{\arraystretch}{2.0}
\caption{Comparison on model size and FPS with the state-of-the-art}
\begin{center}
\begin{tabular}{c|c|c}
\hline
\textbf{\hspace{0.25cm} Model \hspace{0.25cm}}                           & \textbf{\hspace{0.25cm}\#Param (M)\hspace{0.25cm} }  & \textbf{\hspace{0.05cm}\#Frame Per Second\hspace{0.05cm}}\\
\hline
SAFA~\cite{Shi2019}                 & 29.50                                              & 35.90   \\
\hline
Shi et al.~\cite{Shi2020CVPR}  & 17.90                                              & 22.72    \\ 
\hline
EgoTR~\cite{Yang2021}            & 195.90                                            &13.37   \\
\hline
TransGCNN (ours)                    & 87.80                                               & 23.09  \\
\hline
\end{tabular}
\label{table:speed}
\end{center}
\end{table}


\textbf{Training Cost.} Our TransGCNN and the second-performing EgoTR both leverage Transformer-style modules to enhance feature representation learning. However, these two models utilize Transformer in different manners. Specifically, EgoTR employs vision Transformer ViT as backbone to extract the global representation, while our TransGCNN treats the multi-scale window Transformer as the spatial attention module to select salient CNN features as the final embedded features. In order to explore how the incorporation of Transformer-style modules affects the learning process, we depict the learning curves of the two models on CVUSA dataset in Fig.~\ref{fig_learning}~(a). Here, the batch size of 32 is selected to guarantee that EgoTR achieves the best performance.  We observe that both TransGCNN and EgoTR take about 150 epochs to converge. Furthermore, we would also like to know the impacts of batch size on both models, as it is an important hyper-parameter in deep models.  In our practice, we reduce the batch size from 32 down to 16 and 8, and re-train both models on CVUSA dataset. The learning curves of both models with different batch sizes are depicted in Fig.~\ref{fig_learning} (a). Besides, the recall curves of the learned models with different batch sizes on CVUSA dataset are given in Fig.~\ref{fig_learning} (b). We note some particular results from these curves. First, reducing batch size causes a slight decrease in the converge speed of EgoTR. Second, EgoTR suffers from obvious accuracy degradation when the batch size is reduced. In specific, the r@1 drops from 93.69 to 91.02 when the batch size is reduced down to 16, and from 93.69 to 86.49 when the batch size is reduced down to 8. In contrast, our TransGCNN does not show obvious performance degradation when the batch size is reduced. The above observations indicate that the performance of EgoTR is susceptible to batch size, and poses a higher requirement on the video memory of the computing machine compared to our TransGCNN.    

\subsection{Ablation Studies}

\textbf{Does Part-level Transformer Help?}  In TransGCNN, we design a novel multi-scale window Transformer as the spatial attention module to select salient CNN features. Our multi-scale window Transformer consists of a regular global Transformer and a novel part-level Transformer. To show the contributions of the part-level Transformer, we consider two scenarios: (a) TransGCNN-SB, which removes part-level Transformer while increasing the depth of the global Transformer from 2 to 4 in order to keep the amount of parameters the same as the original multi-scale window Transformer; (b) TransGCNN-SS, which replaces the part-level Transformer with the global Transformer for constructing a two-branch single-scale window Transformer. We test the two modified models on CVUSA dataset, and show their recalls in Table~\ref{tab:attention}. We notice obvious accuracy degradation in r@1 (1.46\% and 5.65\% respectively for TransGCNN-SB and TransGCNN-SS), demonstrating the effectiveness of the part-level Transformer in enhancing feature representation learning. This is because the part-level Transformer excels at capturing small-scale information by limiting the calculation of self-attention to a local window. Such local information complements the large-scale information from the global Transformer to obtain more comprehensive information.  
 
\begin{table}[htbp]
\renewcommand{\arraystretch}{2.0}
\caption{Ablation studies of TransGCNN on part-level Transformer}
\begin{center}
\begin{tabular}{c|c|c|c|c}
\hline
\multirow{2}*{\textbf{Model}}    &  \multicolumn{4} {|c} {\textbf{CVUSA}}\\
\cline{2-5} 
~ & \hspace{0.15cm}\textbf{r@1}\hspace{0.15cm} & \hspace{0.15cm}\textbf{r@5}\hspace{0.15cm} & \hspace{0.15cm}\textbf{r@10}\hspace{0.15cm} & \hspace{0.15cm}\textbf{r@1\%}\hspace{0.15cm}\\
\hline
TransGCNN                          & 94.15 & 98.21 & 98.94 & 99.19\\
\hline
TransGCNN-SB                    & 92.69 & 97.66 & 98.58 & 99.71 \\
\hline
TransGCNN-SS                    & 88.50 & 96.81 & 98.13 & 99.71 \\
\hline
w/o SE                                 & 92.96 & 98.01 & 98.91 & 99.72 \\
\hline
\end{tabular}
\label{tab:attention}
\end{center}
\end{table}

\textbf{Does Channel Attention Matter?}  Given feature maps respectively from the global and part-level Transformers, we first perform pixel-wise summation and then apply SE-based channel attention to re-weigh different channels for achieving multi-scale information fusion.  We show the contributions of the channel attention by simply removing the SE block from the multi-scale window Transformer and testing the reduced model on CVUSA dataset. The results are shown in Table~\ref{tab:attention}. We observe that with the help of channel attention, TransGCNN improves by the margin of 1.19\% at r@1 (from 92.96 to 94.15). This observation demonstrates that this dynamic information fusing mechanism does play an important role in our multi-scale window Transformer module. This might be explained by the fact that the SE block helps emphasize informative channels and thus reduce channel redundancy.

\textbf{The Number of Vertical Parts $\bm{w}$.} Intuitively, $w$ determines the granularity of part-level  representations. We investigate the effects of the number of vertical parts $w$ on retrieval accuracy by increasing $w$ from 1, 3, 5 up to 7, and reporting the recall rates under different values of $w$ in Table~\ref{table3}. The results indicate that as $w$ increases from 1 to 5, we obtain better recall performances. Specifically, obvious improvements (2.25\% and 1.45\%) for r@1 are obtained when $w$ increases from 1 to 3, and from 3 to 5. However, the retrieval accuracy does not always increase with $w$. When $w$ increases from 5 up to 7, parts are too small to be discriminative, leading to accuracy degradation. That means, an over-increased $w$ actually compromises the discriminative power of part-level representations. Therefore, we set $w$ to be 5 across all our experiments.  This is reasonable and could be explained by the fact that panoramic street-view images are captured when the car is driving on the road and thus these panoramic images typically consist of five semantic components, including three lawn scenes and two roads. 

\begin{table}[htbp]
\renewcommand{\arraystretch}{2.0}
\caption{Impact of the number of parts $w$ on recall accuracy }
\begin{center}
\begin{tabular}{|c|c|c|c|c|c|}
\hline
\multirow{2}*{\textbf{Number of Parts}}    &  \multicolumn{4} {|c|} {\textbf{CVUSA}}\\
\cline{2-5} 
~ & \hspace{0.15cm}\textbf{r@1}\hspace{0.15cm} & \hspace{0.15cm}\textbf{r@5}\hspace{0.15cm} & \hspace{0.15cm}\textbf{r@10}\hspace{0.15cm} & \hspace{0.15cm}\textbf{r@1\%}\hspace{0.15cm}\\
\hline
$w$ = 1   & 90.35  & 97.25  &  98.52 &  99.79  \\
\hline
$w$ = 3   & 92.60  & 98.01  &  98.72 &  99.78   \\
\hline
$w$ = 5   & \textbf{94.15}   & \textbf{98.21} & \textbf{98.94} & \textbf{99.79} \\
\hline
$w$= 7   & 92.52  & 98.00. &  98.77  &  99.77 \\
\hline
\end{tabular}
\label{table3}
\end{center}
\end{table}

\textbf{The Number of Attention Channels K.} In our work, multiple attention maps are introduced to focus the deep model on salient CNN features, with different attention map capturing different discriminative cues. We investigate the effects of the number of attention channels K on retrieval accuracy by increasing K from 1, 2, 4 up to 8, and report the recall rates under different values of K in Table~\ref{table4}. The results indicate that as K increases, we can obtain better recall performances. Note that significant improvement ($7.37\%$) for r@1 is obtained when K increases from 1 to 2. However, when K increases from 2 to 4 and from 4 to 8, we only obtain improvements in r@1 smaller than $1.5\%$. Therefore, we do not increase K to an even larger number, and set K to be 8 across all experiments.

\begin{table}[htbp]
\renewcommand{\arraystretch}{2.0}
\caption{Impact of Attention Channels K on recall accuracy }
\begin{center}
\begin{tabular}{|c|c|c|c|c|c|}
\hline
\multirow{2}*{\textbf{\#Attention Channels}}    &  \multicolumn{4} {|c|} {\textbf{CVUSA}}\\
\cline{2-5} 
~ & \hspace{0.15cm}\textbf{r@1}\hspace{0.15cm} & \hspace{0.15cm}\textbf{r@5}\hspace{0.15cm} & \hspace{0.15cm}\textbf{r@10}\hspace{0.15cm} & \hspace{0.15cm}\textbf{r@1\%}\hspace{0.15cm}\\
\hline
K = 1       & 84.42  & 95.38  & 97.25 &  99.53  \\
\hline
K = 2       & 91.79  & 95.57  & 98.41 &  99.68   \\
\hline
K = 4       & 92.75  & 98.04  & 98.75 &  99.77 \\
\hline
K = 8       & \textbf{94.15}  & \textbf{98.21} &\textbf{98.94}  & \textbf{99.79} \\
\hline
\end{tabular}
\label{table4}
\end{center}
\end{table}

\begin{figure*}[htbp]
	\centering
	\begin{minipage}{0.48\linewidth}
		\centerline{\includegraphics[width=\textwidth,  height = 0.8in]{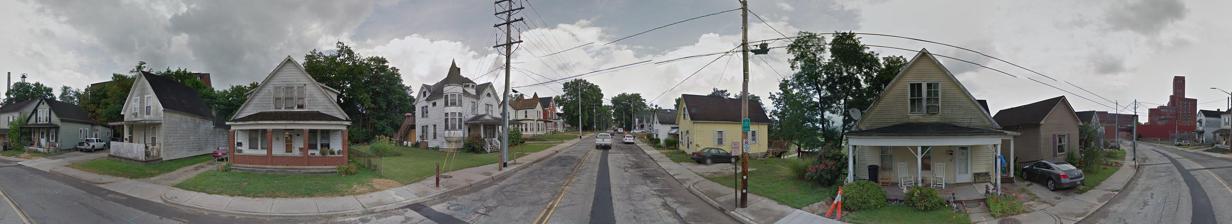}}
	\end{minipage}
	\begin{minipage}{0.48\linewidth}
		\centerline{\includegraphics[width=\textwidth,  height = 0.8in]{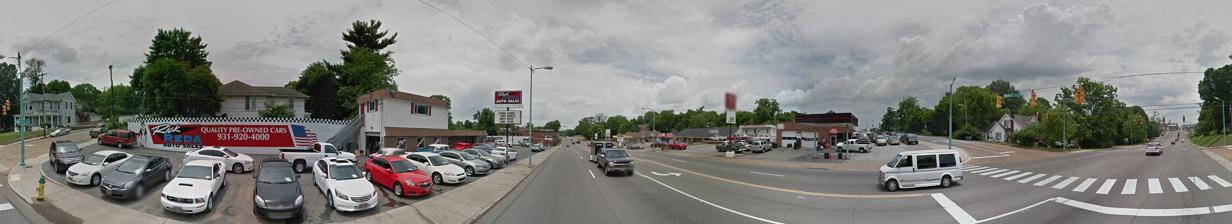}}
	\end{minipage}\vspace{6pt}
        	\centerline{(a) Street-view ground images}
		
\vspace{8pt}
	\begin{minipage}{0.47\linewidth}
		\centerline{\includegraphics[width=\textwidth, height = 0.8in]{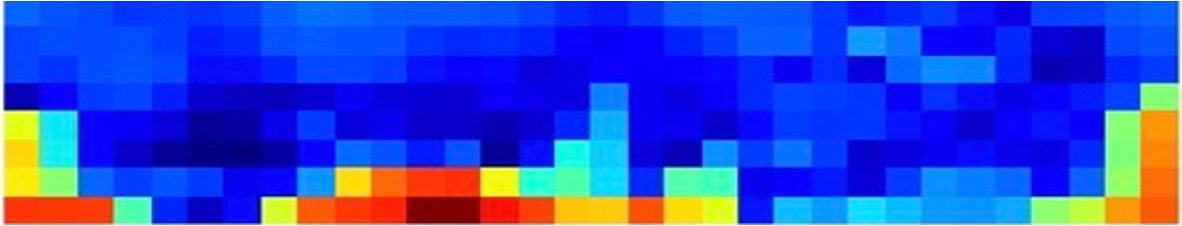}}		
	\end{minipage}
	\begin{minipage}{0.47\linewidth}
		\centerline{\includegraphics[width=\textwidth, height = 0.8in]{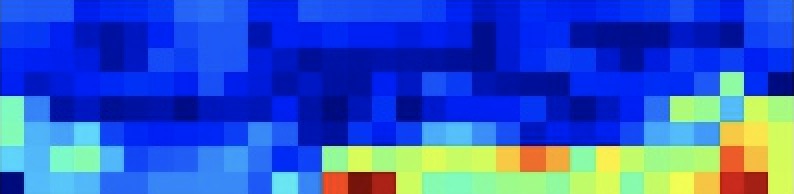}}
	\end{minipage}
	\begin{minipage}{0.02\linewidth}
		\centerline{\includegraphics[width=\textwidth, height = 0.8in]{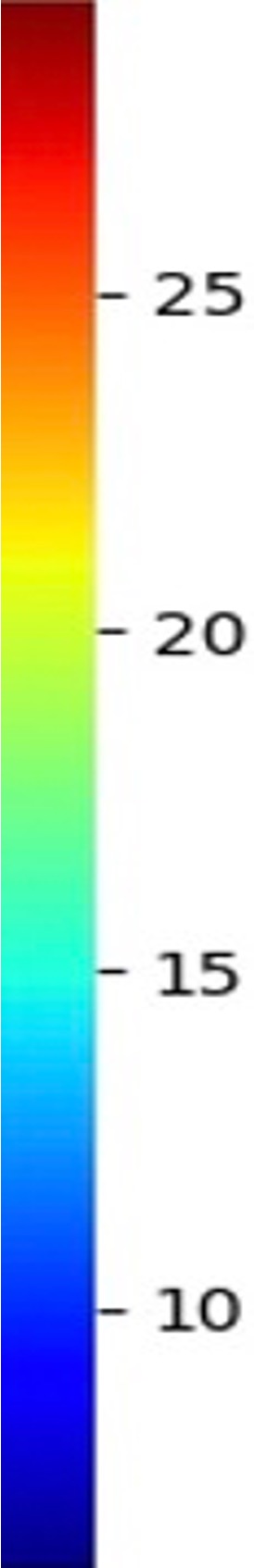}}
	\end{minipage}\vspace{6pt}
	\centerline{(b) Global spatial attention maps}
	
\vspace{8pt}
	\begin{minipage}{0.47\linewidth}
		\centerline{\includegraphics[width=\textwidth, height = 0.8in]{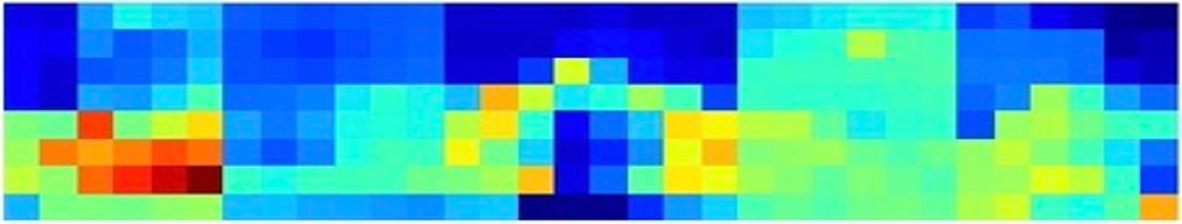}}
	\end{minipage}
	\begin{minipage}{0.47\linewidth}
		\centerline{\includegraphics[width=\textwidth, height = 0.8in]{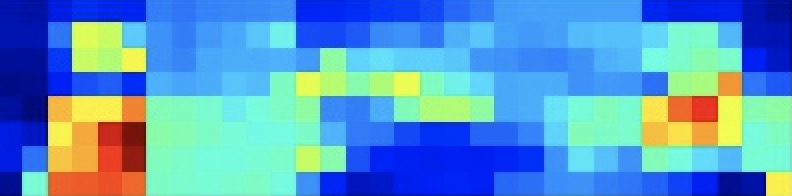}}
	\end{minipage}
	\begin{minipage}{0.02\linewidth}
		\centerline{\includegraphics[width=\textwidth, height = 0.8in]{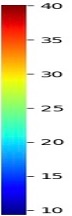}}
	\end{minipage}\vspace{6pt}
	\centerline{(c) Part-level spatial attention maps}
	
\vspace{8pt}
	\begin{minipage}{0.47\linewidth}
		\centerline{\includegraphics[width=\textwidth, height = 0.8in]{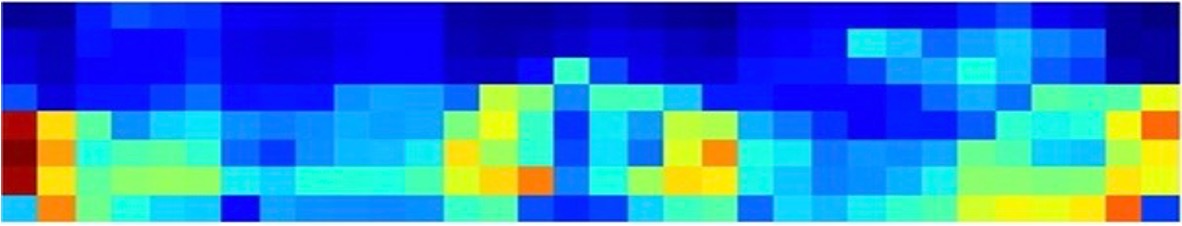}}
	\end{minipage}
	\begin{minipage}{0.47\linewidth}
		\centerline{\includegraphics[width=\textwidth, height = 0.8in]{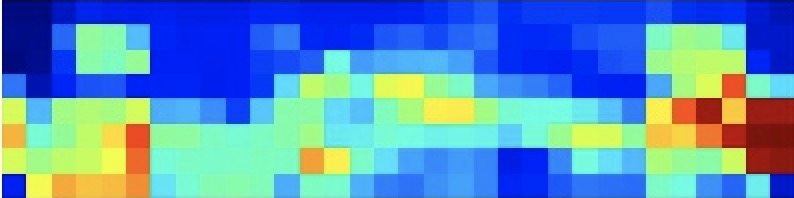}}
	\end{minipage}
	\begin{minipage}{0.02\linewidth}
		\centerline{\includegraphics[width=\textwidth, height = 0.8in]{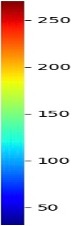}}
	\end{minipage}\vspace{6pt}
	\centerline{(d) Spatial attention maps from our TransGCNN}
	
\vspace{8pt}
	\begin{minipage}{0.47\linewidth}
		\centerline{\includegraphics[width=\textwidth, height = 0.8in]{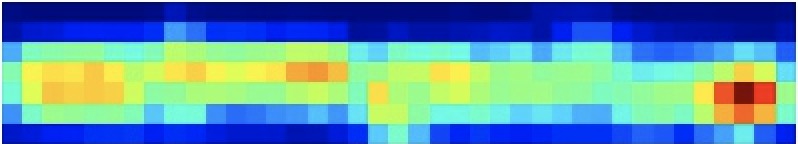}}
	\end{minipage}
	\begin{minipage}{0.47\linewidth}
		\centerline{\includegraphics[width=\textwidth, height = 0.8in]{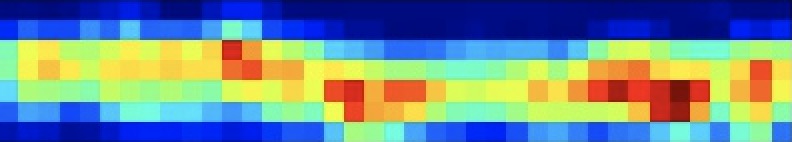}}
	\end{minipage}
	\begin{minipage}{0.02\linewidth}
		\centerline{\includegraphics[width=\textwidth, height = 0.8in]{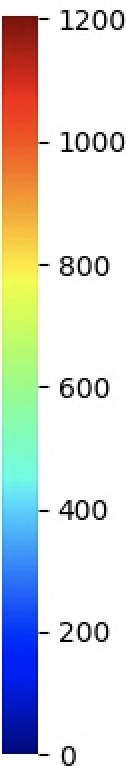}}
	\end{minipage}\vspace{6pt}
	\centerline{(e) Spatial attention maps from the baseline SAFA}

	\caption{Visualizations of spatial attention maps from our TransGCNN and the baseline SAFA.}
	\label{fig_attention}
	
\end{figure*}

\textbf{Visualization of Spatial Attention Maps.} We visualize the spatial attention maps learned from the baseline SAFA and our TransGCNN in Fig.~\ref{fig_attention}. In addition, we also present the enhanced feature maps from both the global and part-level Transformers to illustrate how our TransGCNN actually work. We make the following two observations from these attention maps. First, our TransGCNN is capable of identifying salient roads and buildings, and paying more attention to these objects. In comparison, the baseline SAFA focuses more on distinguishing foreground from background (i.e., sky) and fails to identify the relative importances of different foreground objects. The advantages of our TransGCNN over SAFA further demonstrate the importances of utilizing Transformer-style modules to capture global contexts for implementing spatial attention mechanism.  Second, the global and part-level branches in our proposed multi-scale window Transformer module learns different features, which complement each other to obtain more abundant descriptions of the scenes. 

\begin{figure*}
    \centerline{\includegraphics[scale=0.37]{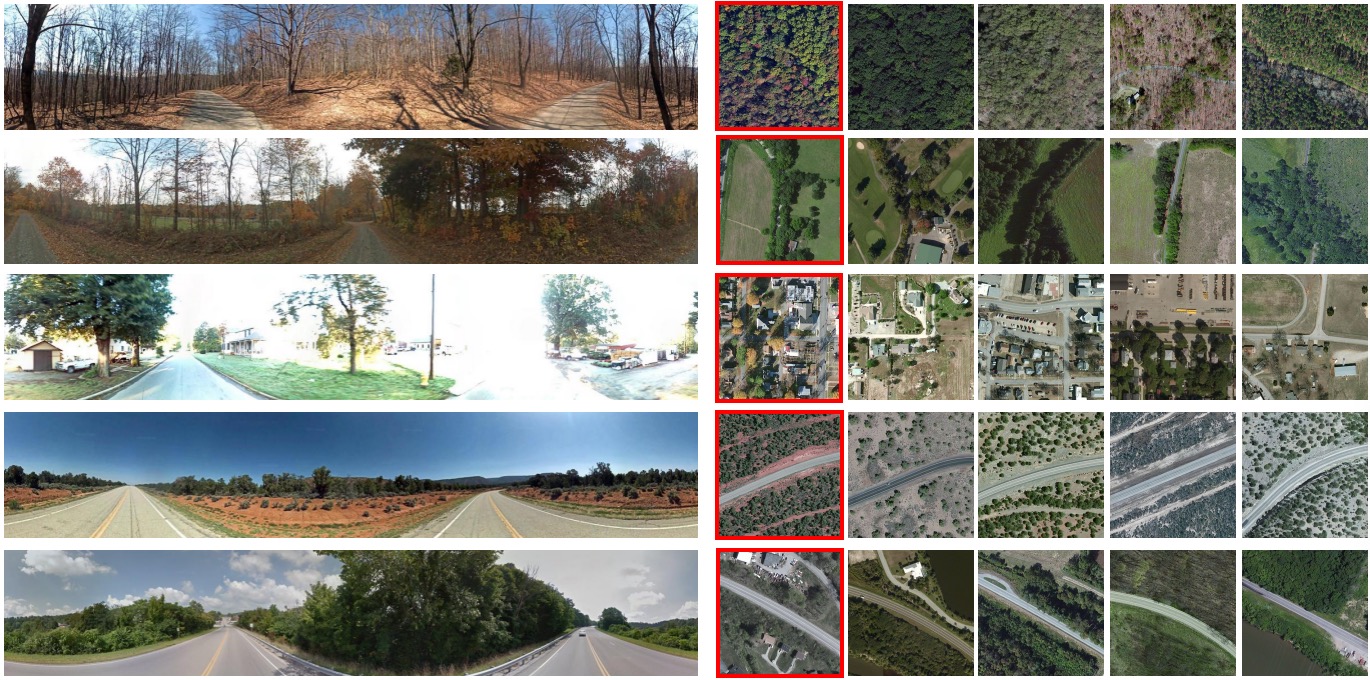}}
    \caption{Image retrieval examples on CVUSA dataset \cite{Zhai2017}. From left to right: street-view query image and the Top 1-5 retrieved satellite images. Red borders indicate ground-truth satellite images.}
    \label{fig8}
    \end{figure*}

\begin{figure*}
    \centerline{\includegraphics[scale=0.37]{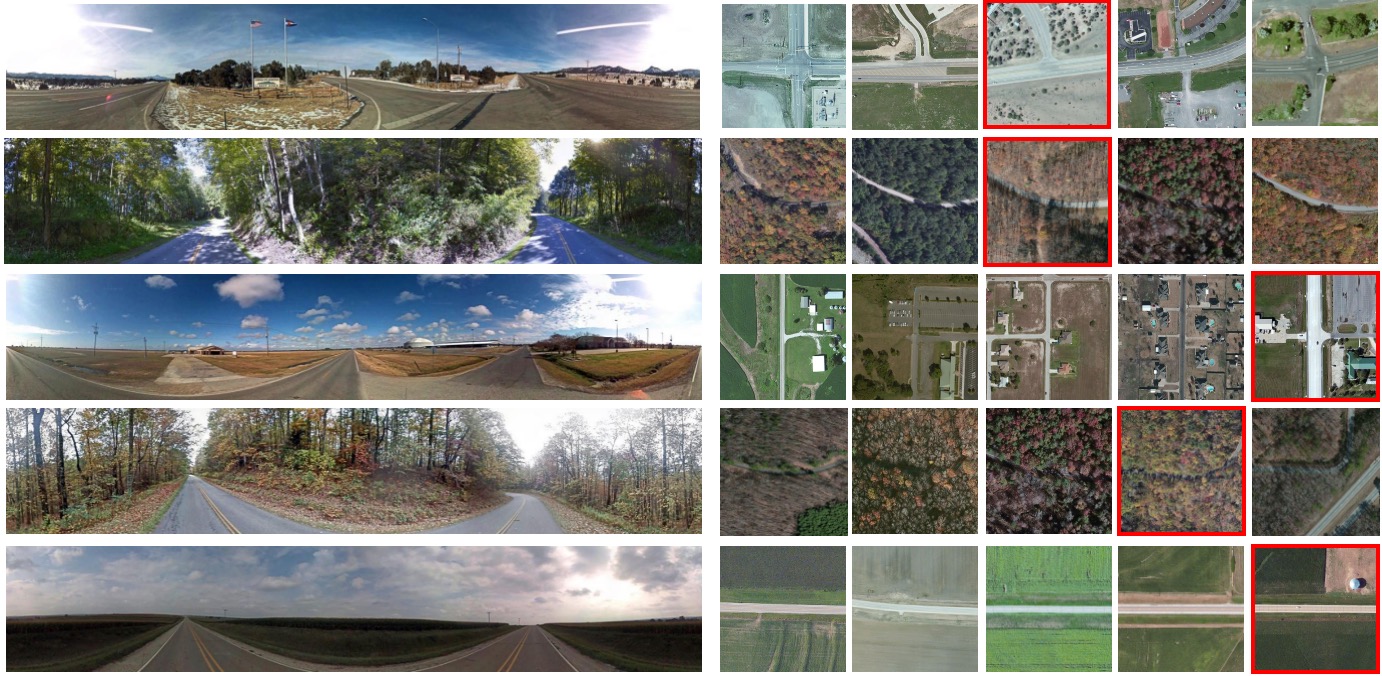}}
    \caption{Examples of negative scenarios, where the ground-truth satellite image is not the nearest top-1 retrieved images according to Euclidean distance. From left to right: street-view query image and the Top 1-5 retrieved satellite image. Red borders indicate ground-truth satellite images.}
    \label{fig9}
    \end{figure*}
    
\subsection{Presentation and Discussion} 
We randomly choose five panoramic street-view images from test dataset and perform image retrieval using our TransGCNN. The top-5 retrieved satellite images are presented in Fig.~\ref{fig8}. We observe that our method could successfully localize ground-truth satellite images even when there exist huge visual differences between ground-to-aerial image pairs.  In addition, we also present some negative scenarios in Fig.~\ref{fig9}, where the ground-truth satellite image is not the nearest top-1 retrieved image according to Euclidean distance in the feature embedding space.  It is easily observed that these negative samples are really hard samples, which are even difficult for human to identify.

\section{Conclusions}
\label{sec:conclusion}

In this paper, we propose a novel Transformer-guided convolutional neural network architecture to couples CNN-based local features with Transformer-based global representations for enhanced representation learning. Specifically, we design a multi-scale window Transformer and treat it as the spatial attention module to select salient CNN features as the final embedded features. In particular, the multi-scale window Transformer is achieved by combining image features from multi-scale windows to produce stronger global contexts.  Extensive experiments on popular benchmark datasets demonstrate that our model achieves top-1 accuracy of 94.12\% and 84.92\% on CVUSA and CVACT\_val, respectively, which outperforms the second-performing baseline with less than 50\% parameters and almost $\mathbf{2\times}$ higher frame rate, thus achieving a preferable accuracy-efficiency tradeoff. In future, we will explore to utilize our proposed method to address the more challenging task of unaligned ground-to-aerial geolocalization.

\section*{Acknowledgment}
This work was supported by in part by National Natural Science Foundation of China under Grant 61921004, and in part by ZhiShan Scholar Program of Southeast University.

\bibliographystyle{IEEEtran}
\bibliography{IEEEabrv,mylib}

\begin{thebibliography}{10}
\providecommand{\url}[1]{#1}
\csname url@samestyle\endcsname
\providecommand{\newblock}{\relax}
\providecommand{\bibinfo}[2]{#2}
\providecommand{\BIBentrySTDinterwordspacing}{\spaceskip=0pt\relax}
\providecommand{\BIBentryALTinterwordstretchfactor}{4}
\providecommand{\BIBentryALTinterwordspacing}{\spaceskip=\fontdimen2\font plus
\BIBentryALTinterwordstretchfactor\fontdimen3\font minus
  \fontdimen4\font\relax}
\providecommand{\BIBforeignlanguage}[2]{{%
\expandafter\ifx\csname l@#1\endcsname\relax
\typeout{** WARNING: IEEEtran.bst: No hyphenation pattern has been}%
\typeout{** loaded for the language `#1'. Using the pattern for}%
\typeout{** the default language instead.}%
\else
\language=\csname l@#1\endcsname
\fi
#2}}
\providecommand{\BIBdecl}{\relax}
\BIBdecl

\bibitem{Senlet2012}
T.~Senlet and A.~Elgammal, ``Satellite image based precise robot localization
  on sidewalks,'' in \emph{IEEE International Conference on Robotics and
  Automation}.\hskip 1em plus 0.5em minus 0.4em\relax IEEE, 2012, pp.
  2647--2653.

\bibitem{Mcmanus2014}
C.~McManus, W.~Churchill, W.~Maddern, A.~D. Stewart, and P.~Newman, ``Shady
  dealings: Robust, long-term visual localisation using illumination
  invariance,'' in \emph{IEEE international conference on robotics and
  automation (ICRA)}.\hskip 1em plus 0.5em minus 0.4em\relax IEEE, 2014, pp.
  901--906.

\bibitem{Middelberg2014}
S.~Middelberg, T.~Sattler, O.~Untzelmann, and L.~Kobbelt, ``Scalable 6-dof
  localization on mobile devices,'' in \emph{European conference on computer
  vision}.\hskip 1em plus 0.5em minus 0.4em\relax Springer, 2014, pp. 268--283.

\bibitem{Workman2015L}
S.~Workman and N.~Jacobs, ``On the location dependence of convolutional neural
  network features,'' in \emph{IEEE Conference on Computer Vision and Pattern
  Recognition Workshops}, 2015, pp. 70--78.

\bibitem{Workman2015W}
S.~Workman, R.~Souvenir, and N.~Jacobs, ``Wide-area image geolocalization with
  aerial reference imagery,'' in \emph{IEEE International Conference on
  Computer Vision}, 2015, pp. 3961--3969.

\bibitem{Lin2015}
T.-Y. Lin, Y.~Cui, S.~Belongie, and J.~Hays, ``Learning deep representations
  for ground-to-aerial geolocalization,'' in \emph{IEEE conference on Computer
  Vision and Pattern Recognition}, 2015, pp. 5007--5015.

\bibitem{Vo2016}
N.~Vo and J.~Hays, ``Localizing and orienting street views using overhead
  imagery,'' in \emph{European conference on computer vision}.\hskip 1em plus
  0.5em minus 0.4em\relax Springer, 2016, pp. 494--509.

\bibitem{Vo2017}
N.~Vo, N.~Jacobs, and J.~Hays, ``Revisiting im2gps in the deep learning era,''
  in \emph{IEEE International Conference on Computer Vision}, 2017, pp.
  2621--2630.

\bibitem{Hu2018}
S.~Hu, M.~Feng, R.~M. Nguyen, and G.~H. Lee, ``Cvm-net: Cross-view matching
  network for image-based ground-to-aerial geo-localization,'' in \emph{IEEE
  Conference on Computer Vision and Pattern Recognition}, 2018, pp. 7258--7267.

\bibitem{Sun2019}
B.~Sun, C.~Chen, Y.~Zhu, and J.~Jiang, ``Geocapsnet: ground to aerial view
  image geo-localization using capsule network,'' in \emph{IEEE Conference on
  Multimedia and Expo (ICME)}, 2019, pp. 742--747.

\bibitem{Cai2019}
S.~Cai, Y.~Guo, S.~Khan, J.~Hu, and G.~Wen, ``Ground-to-aerial image
  geo-localization with a hard exemplar reweighting triplet loss,'' in
  \emph{IEEE/CVF International Conference on Computer Vision}, 2019, pp.
  8391--8400.

\bibitem{Shi2019}
Y.~Shi, L.~Liu, X.~Yu, and H.~Li, ``Spatial-aware feature aggregation for
  cross-view image based geo-localization,'' in \emph{33rd Conference on Neural
  Information and Processing Systems (NIPS)}, 2019.

\bibitem{Vaswani2017}
A.~Vaswani, N.~Shazeer, N.~Parmar, J.~Uszkoreit, L.~Jones, A.~Gomze, L.~Kaiser,
  and I.~Polosukhin, ``Attention is all you need,'' in \emph{Advances in Neural
  Information Processing Systems}, 2017.

\bibitem{Yang2021}
H.~Yang, X.~Lu, and Y.~Zhu, ``Cross-view geo-localization with evolving
  transformer,'' in \emph{35th Conference on Neural Information and Processing
  Systems (NIPS)}, 2021.

\bibitem{Dosovitskiy2021}
A.~Dosovitskiy, L.~Beyer, A.~Kolesnikov, D.~Weissenborn, and et~al., ``An image
  is worth 16x16 words: Transformers for image recognition at scale,'' in
  \emph{International Conference on Learning Representations}, 2021.

\bibitem{Simonyan2015}
K.~Simonyan and A.~Zisserman, ``Very deep convolutional networks for
  large-scale image recognition,'' in \emph{International Conference on
  Learning Representations}, 2015.

\bibitem{Zhai2017}
M.~Zhai, Z.~Bessinger, S.~Workman, and N.~Jacobs, ``Predicting ground-level
  scene layout from aerial imagery,'' in \emph{IEEE Conference on Computer
  Vision and Pattern Recognition}, 2017, pp. 867--875.

\bibitem{Liu2019}
L.~Liu and H.~Li, ``Lending orientation to neural networks for cross-view
  geolocalization,'' in \emph{IEEE Conference on Computer Vision and Pattern
  Recognition}, 2019.

\bibitem{Hays2008}
J.~Hays and A.~A. Efros, ``Im2gps: estimating geographic information from a
  single image,'' in \emph{IEEE Conference on Computer Vision and Pattern
  Recognition}.\hskip 1em plus 0.5em minus 0.4em\relax IEEE, 2008, pp. 1--8.

\bibitem{Lin2013}
T.-Y. Lin, S.~Belongie, and J.~Hays, ``Cross-view image geolocalization,'' in
  \emph{IEEE Conference on Computer Vision and Pattern Recognition}, 2013, pp.
  891--898.

\bibitem{Dalal2005}
N.~Dalal and B.~Triggs, ``Histograms of oriented gradients for human
  detection,'' in \emph{IEEE Conference on Computer Vision and Pattern
  Recognition}, vol.~1.\hskip 1em plus 0.5em minus 0.4em\relax IEEE, 2005, pp.
  886--893.

\bibitem{Shechtman2007}
E.~Shechtman and M.~Irani, ``Matching local self-similarities across images and
  videos,'' in \emph{IEEE Conference on Computer Vision and Pattern
  Recognition}.\hskip 1em plus 0.5em minus 0.4em\relax IEEE, 2007, pp. 1--8.

\bibitem{Oliva2001}
A.~Oliva and A.~Torralba, ``Modeling the shape of the scene: A holistic
  representation of the spatial envelope,'' \emph{International journal of
  computer vision}, vol.~42, no.~3, pp. 145--175, 2001.

\bibitem{lowe2004}
D.~G. Lowe, ``Distinctive image features from scale-invariant keypoints,''
  \emph{International journal of computer vision}, vol.~60, no.~2, pp. 91--110,
  2004.

\bibitem{Bay2006}
H.~Bay, T.~Tuytelaars, and L.~Van~Gool, ``Surf: Speeded up robust features,''
  in \emph{European Conference on Computer Vision}.\hskip 1em plus 0.5em minus
  0.4em\relax Springer, 2006, pp. 404--417.

\bibitem{Rublee2011}
E.~Rublee, V.~Rabaud, K.~Konolige, and G.~Bradski, ``Orb: An efficient
  alternative to sift or surf,'' in \emph{International Conference on Computer
  Vision}.\hskip 1em plus 0.5em minus 0.4em\relax IEEE, 2011, pp. 2564--2571.

\bibitem{Bastanlar2010}
Y.~Bastanlar, A.~Temizel, and Y.~Yardimci, ``Improved sift matching for image
  pairs with scale difference,'' \emph{Electronics Letters}, vol.~46, no.~5,
  pp. 346--348, 2010.

\bibitem{Huijuan2011}
Z.~Huijuan and H.~Qiong, ``Fast image matching based-on improved surf
  algorithm,'' in \emph{International conference on electronics, communications
  and control (ICECC)}.\hskip 1em plus 0.5em minus 0.4em\relax IEEE, 2011, pp.
  1460--1463.

\bibitem{Sedaghat2015}
A.~Sedaghat and H.~Ebadi, ``Remote sensing image matching based on adaptive
  binning sift descriptor,'' \emph{IEEE transactions on geoscience and remote
  sensing}, vol.~53, no.~10, pp. 5283--5293, 2015.

\bibitem{Karami2017}
E.~Karami, S.~Prasad, and M.~Shehata, ``Image matching using sift, surf, brief
  and orb: performance comparison for distorted images,'' \emph{arXiv preprint
  arXiv:1710.02726}, 2017.

\bibitem{Lefevre2017}
S.~Lef{\`e}vre, D.~Tuia, J.~D. Wegner, T.~Produit, and A.~S. Nassar, ``Toward
  seamless multiview scene analysis from satellite to street level,''
  \emph{Proceedings of the IEEE}, vol. 105, no.~10, pp. 1884--1899, 2017.

\bibitem{Noda2010}
M.~Noda, T.~Takahashi, D.~Deguchi, I.~Ide, H.~Murase, Y.~Kojima, and T.~Naito,
  ``Vehicle ego-localization by matching in-vehicle camera images to an aerial
  image,'' in \emph{Asian Conference on Computer Vision}.\hskip 1em plus 0.5em
  minus 0.4em\relax Springer, 2010, pp. 163--173.

\bibitem{Viswanathan2014}
A.~Viswanathan, B.~R. Pires, and D.~Huber, ``Vision based robot localization by
  ground to satellite matching in gps-denied situations,'' in \emph{IEEE/RSJ
  International Conference on Intelligent Robots and Systems}.\hskip 1em plus
  0.5em minus 0.4em\relax IEEE, 2014, pp. 192--198.

\bibitem{Arandjelovic2016}
R.~Arandjelovic, P.~Gronat, A.~Torii, T.~Pajdla, and J.~Sivic, ``Netvlad: Cnn
  architecture for weakly supervised place recognition,'' in \emph{IEEE
  conference on computer vision and pattern recognition}, 2016, pp. 5297--5307.

\bibitem{He2016}
K.~He, X.~Zhang, S.~Ren, and J.~Sun, ``Deep residual learning for image
  recognition,'' in \emph{IEEE conference on computer vision and pattern
  recognition}, 2016, pp. 770--778.

\bibitem{Sabour2017}
S.~Sabour, N.~Frosst, and G.~Hinton, ``Dynamic routing between capsules,'' in
  \emph{IEEE Conference on Neural Information Processing Systems}, 2017, pp.
  3859--3869.

\bibitem{Hu2018SENet}
J.~Hu, L.~Shen, S.~Albanie, G.~Sun, and E.~Wu, ``Squeeze-and-excitation
  network,'' in \emph{IEEE/CVF Conference on Computer Vision and Pattern
  Recognition}, 2018.

\bibitem{Wang2021}
T.~Wang, Z.~Zheng, C.~Yan \emph{et~al.}, ``Each part matters: Local patterns
  facilitate cross-view geo-localization,'' in \emph{arXiv:2008.11646}, 2021.

\bibitem{Regmi2019}
K.~Regmi and M.~Shah, ``Bridging the domain gap for ground-to-aerial image
  matching,'' in \emph{IEEE/CVF International Conference on Computer Vision},
  2019, pp. 470--479.

\bibitem{Shi2020}
Y.~Shi, X.~Yu, L.~Liu, T.~Zhang, and H.~Li, ``Optimal feature transport for
  cross-view image geo-localization,'' in \emph{AAAI Conference on Artificial
  Intelligence}, 2020, pp. 11\,990--11\,997.

\bibitem{Shi2020CVPR}
Y.~Shi, X.~Yu, D.~Campbell, and H.~Li, ``Where am i looking at? joint location
  and orientation estimation by cross-view matching,'' in \emph{IEEE/CVF
  Conference on Computer Vision and Pattern Recognition}, 2020, pp. 4064--4072.

\end{thebibliography}

\end{document}